\documentclass[letterpaper, 10 pt, conference]{ieeeconf}  

\IEEEoverridecommandlockouts                              

\usepackage[bookmarks=true]{hyperref}
\usepackage{amsmath} 
\usepackage{multicol}
\usepackage{multirow}
\usepackage{todonotes}
\usepackage{amssymb}
\usepackage{array}
\usepackage{empheq}
\usepackage[caption=false,font=footnotesize]{subfig}
\usepackage[noadjust]{cite}
\usepackage{color}
\usepackage[export]{adjustbox}
\usepackage[noadjust]{cite}
\usepackage{graphicx} 

\usepackage{comment}

\newcommand{\tosay}{true}
\newcommand{\ludo}[1]{
{\ifthenelse{\boolean{\tosay}}{\begin{quotation}\textcolor{blue}{Ludo: #1}\end{quotation}}{}}}

\makeatletter
\newcommand{\thickhline}{%
    \noalign {\ifnum 0=`}\fi \hrule height 1.5pt
    \futurelet \reserved@a \@xhline
}
\newcolumntype{"}{@{\hskip\tabcolsep\vrule width 1pt\hskip\tabcolsep}}
\makeatother

\title{\LARGE \bf
    Efficient Humanoid Contact Planning using Learned Centroidal Dynamics Prediction
}

\author{
    Yu-Chi Lin$^{1}$, Brahayam Ponton$^{2,3}$, Ludovic Righetti$^{3,4}$, and Dmitry Berenson$^{1}$
    \thanks{$^{1}$University of Michigan, Ann Arbor, MI, USA, $^{2}$University of Tuebingen, Tuebingen, Germany, $^{3}$Max Planck Institute for Intelligent Systems, Tuebingen, Germany, $^{4}$New York University, New York, NY, USA}%
    \thanks{Part of this work was supported by New York University, the Max-Planck Society, the European Union’s Horizon 2020 research and innovation program (grant agreement No 780684 and European Research Council’s grant No 637935), the National Science Foundation under grant CMMI-1825993, and the Office of Naval Research under grant N000141712050.}
}

\renewcommand{\baselinestretch}{.97}

\begin{document}

\maketitle
\thispagestyle{empty}
\pagestyle{empty}

\begin{abstract}

Humanoid robots dynamically navigate an environment by interacting with it via contact wrenches exerted at intermittent contact poses. Therefore, it is important to consider dynamics when planning a contact sequence. Traditional contact planning approaches assume a quasi-static balance criterion to reduce the computational challenges of selecting a contact sequence over a rough terrain. This however limits the applicability of the approach when dynamic motions are required, such as when walking down a steep slope or crossing a wide gap. Recent methods overcome this limitation with the help of efficient mixed integer convex programming solvers capable of synthesizing dynamic contact sequences. Nevertheless, its exponential-time complexity limits its applicability to short time horizon contact sequences within small environments. In this paper, we go beyond current approaches by learning a prediction of the dynamic evolution of the robot centroidal momenta, which can then be used for quickly generating dynamically robust contact sequences for robots with arms and legs using a search-based contact planner. We demonstrate the efficiency and quality of the results of the proposed approach in a set of dynamically challenging scenarios.

\end{abstract}

\section{Introduction}

\begin{figure}[t]
	\centering
	\includegraphics[scale=0.5]{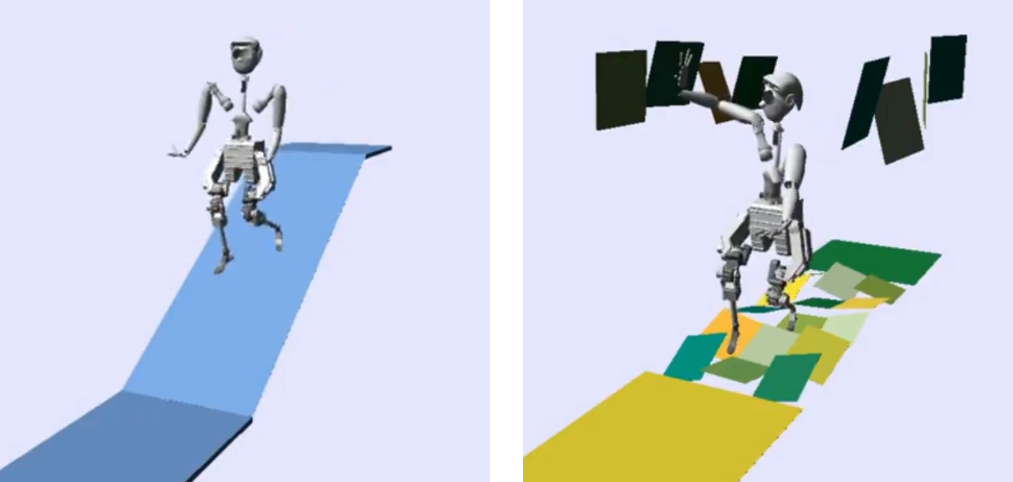}
	\caption{\textcolor{black}{Left: The robot goes down a steep slope where quasi-static motions are not available. Right: The robot goes through a rubble corridor using both palm and foot contacts.}}
	\label{intro_fig}
\end{figure}

Humanoid robots keep balance and navigate uncertain environments by controlling the contact interaction wrenches applied at selected end-effector contact poses. In this work, we are interested in the efficient planning of such sequences of contact poses that can be used by a robot with arms and legs to optimally traverse highly dynamic, large and unstructured environments, as shown in Figure \ref{intro_fig}. Being able to use multiple end-effectors to interact with the environment is beneficial for robot balance and control, but it poses important computational challenges. First of all, the use of multiple non-coplanar contacts prevents the use of simplified dynamic models, such as the linear inverted pendulum model. Second, more computationally demanding balance checks are required because of the non-planar nature of the terrains. Finally, in multi-contact motion scenarios, the robot can use any combination of its available end-effectors, which increases the planning complexity, and further emphasizes the need for fast evaluation of contact pose feasibility.

To cope with the computationally costly planning of dynamically feasible contact sequences, previous approaches trade-off different factors. For instance, on the one hand, some approaches \cite{escande,hauser,reachability_planner_journal,motion_plan_library,CES} use a quasi-static balance criteria \textcolor{black}{\cite{bresler1950forces}}, which lowers computational complexity but does not consider dynamic planning of contacts \cite{bretl_balance,giwc,fast_static_balance}. On the other hand, for more dynamic motions, such as when crossing a wide gap or walking down a steep slope, contact planners based on mixed-integer programming \cite{DBLP:conf/iros/IbanezBP14,DBLP:journals/corr/Aceituno-Cabezas16,MICP_quadruped_2} that can account for dynamics are better suited, but still suffer from the high branching factor of the search, which in large environments still remains computationally demanding for online contacts planning.

In this work, we incorporate motion dynamics within a search-based contact planner. We formulate the contact planning problem as a graph search problem where each edge corresponds to a contact transition, and the motion dynamics are evaluated for each edge. Considering motion dynamics enables the contact planner to not only plan contact sequences for dynamic motions, but also select new contacts based on a measure of ``dynamical robustness'' to achieve robust locomotion. To deal with the computationally heavy optimization of motion dynamics within the contact planning loop, we train neural networks to predict the dynamic evolution of optimal robot momentum over contact transitions, and query the networks in the planning loop to inform the contact planner how to produce contact sequences which are likely to be dynamically-robust. Using a learned approximation of optimal momentum evolution allows us to consider dynamic feasibility of transitions without paying the high computational cost of solving a dynamics optimization problem for each considered edge in the graph. The generated contact sequence is then used by a centroidal momentum dynamics optimizer \cite{centroidal_dynopt_2} to produce a time-optimal dynamically feasible motion plan. To the best of our knowledge, this work is the first attempt where a learned dynamics model is used for online planning of contact sequences for a humanoid robot involving both foot and palm contacts.

In our experiments, we compare our method to a quasi-static search-based and a mixed-integer contact planner. Our results suggest that our approach produces more dynamically robust motions compared to the quasi-static planner which allows us to traverse dynamically challenging environments, and can be orders of magnitude more efficient than mixed-integer based planners in large unstructured environments.

\section{Related Work}

Footstep planning for humanoid robots has been studied extensively \cite{footstep1,footstep2,footstep3,footstep4,footstep5,footstep6,footstep7}. In these works, the planner plans a footstep sequence to avoid obstacles on the ground and remain inside the specified contact regions on a flat or piecewise-flat ground. To increase the likelihood of success, they incorporate an approximation of robot balance and kinematic reachability into the contact transition model, and do not explicitly perform balance check online. There are also works addressing contact planning in unstructured environment using both palm and foot contacts \cite{escande,hauser,reachability_planner_journal,motion_plan_library,CES}. However, these approaches assume quasi-static motions, and drop solutions involving dynamic motions.

Approaches to synthesize dynamically feasible multi-contact motions have also been extensively studied \cite{Herzog-2016b,JustinMomentumOptimization,Dai:2016hz,Caron:2016wt,DBLP:conf/iros/AudrenVKEKY14}. However, it is not trivial to include planning of contact poses in these approaches because contacts planning in general involves discrete or non-convex constraints for the contact poses. \cite{footstep7} addresses the non-convexity by decomposing the environment into a set of convex regions and approximating the rotation using piecewise affine functions. The problem is then formulated as a mixed integer convex program and solved to global optimality. Although \cite{footstep7} only uses foot contact, and does not consider dynamics, it points a direction to include contact planning in an optimization problem.

Extensions of \cite{footstep7} for dynamic planning of a contact sequences are proposed in \cite{DBLP:conf/iros/IbanezBP14, centroidal_dynopt_1}, which extend \cite{footstep7} with the selection of contact timings or hand contacts respectively. More recent works \cite{MICP_quadruped_1,MICP_quadruped_2} use the same concept to plan gait sequences for quadruped robots and produce dynamically robust motions. However, mixed-integer approaches scale poorly against the number of integer decision variables. For instance, their applicability is limited to online contact generation in environments with few convex terrain regions, and short planning horizons.

\cite{kinodynamic_contact_planner} proposes a kinodynamic sampling-based contact planner to plan kinodynamically feasible contact sequences. They use a simplified robot model to dynamically plan smooth center of mass (CoM) trajectories based on convex optimization and then search for kinematically feasible contact poses around it. It shows a unified planning framework to consider dynamics and kinematics constraints, but it suffers from long planning time. \cite{croc} proposes an efficient dynamic feasibility check by conservatively reformulating the problem as a linear program. While the check guarantees to reject dynamically infeasible motions, they do not address dynamical robustness in the stability check. \cite{learned_dynamics_footstep_planning} learns quadratic dynamics objective of humanoid walking motion, and apply this learned model to select steps in a search-based footstep planner. However, their dynamics model assumes flat contact, and does not consider palm contacts, which limits the applicability of the approach.

\section{Problem Statement}

In this paper, we focus our efforts on the dynamic planning of contact sequences for humanoid robots. Given an environment specified as a set of polygonal surfaces, a start stance, and a goal region, we seek to produce a dynamically-feasible contact sequence along with a dynamics sequence, which includes centroidal momentum trajectories and contact wrenches at each time step of the trajectory, to move the robot from the start stance to the goal region within a specified planning time. The robot always uses feet contacts, but can also optionally use palm contacts when they are available. As considering variable transition times significantly increases the branching factor of the search, we assume fixed timing for each contact transition. We also assume the friction coefficient of the environment is given and fixed.

\section{Centroidal Momentum Dynamics Optimization} \label{com_dynopt}
	
The momentum dynamics have been widely adopted to plan dynamically feasible motions for floating base robots \cite{OrinCentroidalMomentum, Kajita:2003gj}. In this work, we use the fixed-time formulation of the centroidal dynamics optimizer proposed in \cite{centroidal_dynopt_2}. In the following, we briefly summarize them and explain how we use them to generate robust motion plans. The dynamics of a floating-base robot with $n$ degrees of freedom is
\begin{equation}
\mathbf{H(q)\ddot{q} + C(q,\dot{q})} = \mathbf{S}^{T}\tau + \mathbf{J}^{T}_{e}\lambda
\label{full_dynamics}
\end{equation}
\noindent where $\mathbf{q}=\left[q^{T}, x^{T}\right]^{T}$ denotes the generalized robot states including joint positions $q \in \mathbb{R}^{n}$, and floating base frame $x \in SE(3)$. $\mathbf{H} \in \mathbb{R}^{(n+6) \times (n+6)}$ is the inertia matrix, and $\mathbf{C} \in \mathbb{R}^{n+6}$ stands for the Coriolis, centrifugal, and gravity forces. $\mathbf{S} = \left[\mathbf{I}^{n\times n} \ 0\right]$ is a selection matrix, $\tau \in \mathbb{R}^{n}$ is the torques vector, $J_{e}$ is the end-effector jacobian, and $\lambda = \left[\cdots \mathbf{f}^{T}_{e} \ \tau^{T}_{e}\cdots\right]^{T}$ comprises the force $\mathbf{f}_{e}$ and torque $\tau_{e}$ of each end-effector contact. We can then decompose Eq. \eqref{full_dynamics} to actuated parts (Eq. \eqref{actuated_dynamics}), and unactuated parts (Eq. \eqref{unactuated_dynamics})
\begin{subequations}
	\begin{empheq}{align}
	\mathbf{H_{a}(q)\ddot{q} + C_{a}(q,\dot{q})} & = \tau + \mathbf{J^{T}_{e,a}}\lambda \label{actuated_dynamics} \\
	\mathbf{H_{u}(q)\ddot{q} + C_{u}(q,\dot{q})} & = \mathbf{J^{T}_{e,u}}\lambda \label{unactuated_dynamics}
	\end{empheq}
\end{subequations}
Under the assumption that enough torque can always be generated by the robot, if there exist robot states $\mathbf{q,\dot{q},\ddot{q}}$, and the external forces $\lambda$ that satisfy Eq. \eqref{unactuated_dynamics}, Eq. \eqref{actuated_dynamics} is also satisfied. With the assumption and decomposition, Eq. \eqref{unactuated_dynamics} verifies the dynamic feasibility, and Eq. \eqref{actuated_dynamics} is only required to verify torque limits and kinematic constraints. Eq. \eqref{unactuated_dynamics} is equivalent to the Newton-Euler equations of the robot \cite{Wieber2006}, which means that the momentum rate equals the applied external contact wrenches. The centroidal dynamics expressed at the robot CoM is
\begin{equation}
\begin{bmatrix}
\mathbf{\dot{r}}\\ 
\mathbf{\dot{l}}\\ 
\mathbf{\dot{k}}
\end{bmatrix} = 
\begin{bmatrix}
\frac{1}{M}\mathbf{l}\\ 
M\mathbf{g} + \sum \mathbf{f}_{e}\\ 
\sum(T_{e}(\mathbf{z}_{e})-\mathbf{r})\times \mathbf{f}_{e} + \tau_{e}
\end{bmatrix}
\label{centroidal_dynamics}
\end{equation}
\noindent $\mathbf{r}$ is the CoM position. $\mathbf{l}$ and $\mathbf{k}$ are linear and angular momenta, respectively. $M$ is the robot mass. $\mathbf{z}_{e}$ is the center of pressure (CoP) of each contact in the contact frame. $\mathbf{f}_{e}$ and $\tau_{e}$ are the contact force and torque at the CoP of each end-effector and finally, $T_{e}$ is a coordinate transform in the CoM frame. In addition to Eq.~\eqref{centroidal_dynamics}, contact forces need to be inside friction cones, and CoPs inside the support regions of each contact, to prevent the contact from sliding and tilting. \textcolor{black}{A contact transition is dynamically feasible if there exists a sequence of centroidal momenta and contact wrenches obeying the above constraints.}
	
To compute a dynamically robust motion we follow \cite{centroidal_dynopt_2} to minimize the weighted sum of the square norm of $\mathbf{l}$, $\mathbf{\dot{l}}$, $\mathbf{k}$, $\mathbf{\dot{k}}$, $\mathbf{f}_{e}$, and $\tau_{e}$. Lower $\mathbf{l}$ and $\mathbf{\dot{l}}$ help improve dynamic stability \cite{safe_locomotion}. Reducing $\mathbf{k}$ and $\mathbf{\dot{k}}$ help the robot perform more natural motion \cite{ang_mom_human_walking}. $\mathbf{f}_{e}$ and $\tau_{e}$ terms encourage a more even distribution of forces and torques over all the contacts, which increase controllability of the robot. Additionally, we append two terms, the lateral contact forces $\mathbf{f}_{l}$ in the contact frame
\begin{equation}
\mathbf{f}_{l} = [\mathbf{f}_{c}[x], \mathbf{f}_{c}[y]]^{T}, \mathbf{f}_{c} = T_{e}^{-1}(\mathbf{f}_{e})
\label{lateral_contact_force}
\end{equation}
\noindent and the weighted CoP position $\mathbf{z}_{w}$ in each contact frame
\begin{equation}
\mathbf{z}_{w} = [\frac{\mathbf{z}_{e}[x]}{l_{e,x}}, \frac{\mathbf{z}_{e}[y]}{l_{e,y}}]^{T}
\label{cop_distance}
\end{equation}
\noindent where $l_{e,x}$ and $l_{e,y}$ are the lengths of the support region in X and Y direction of the contact frame. These two additional terms capture the robustness of the contact. A lower lateral contact forces favor forces away from the friction cone limits and therefore decrease the chances of sliding while a CoP position closer to the contact center decreases chance of contact tilting during execution.
	
Here, the dynamics optimization does not have a CoM position goal and we do not specify the final CoM position as part of the objective. Instead, a final CoM position bound is enforced as a constraint based on the mean position of the last pair of feet contacts. In the final time step of the whole contact sequence, we also constrain the CoM velocity to zero to ensure the robot can finally come to a stop.

\section{Anytime Discrete-Search Contact Planner} \label{anytime_contact_planner}

\begin{figure}[t]
	\centering
	\medskip
	\includegraphics[scale=0.32]{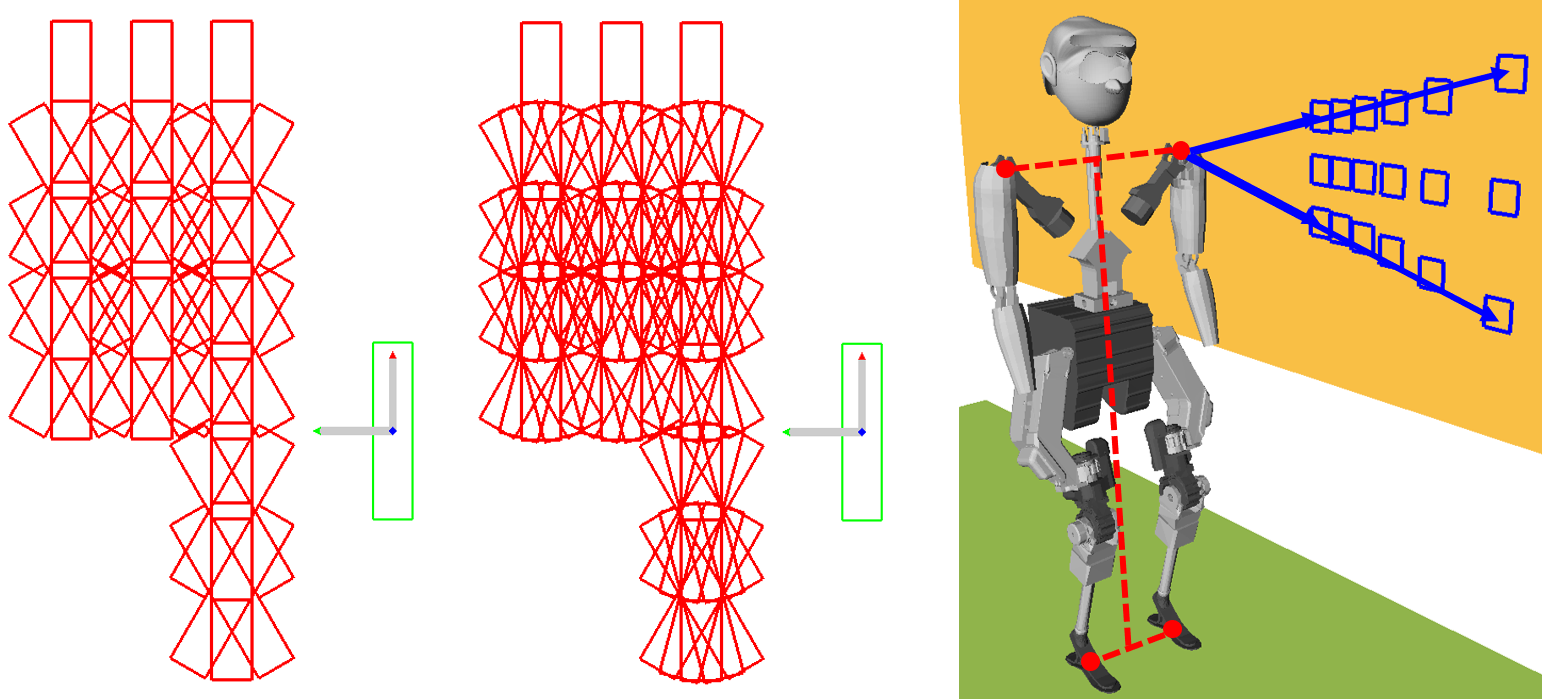}
	\caption{Left: The foot contact transition model used in training data collection. (38 steps) Middle: The foot contact transition model used in the experiment. (60 steps) Right: The palm contact transition model, expressed as the projections from the approximated shoulder to a wall.}
	\label{step_transition_model}
\end{figure}

We build on the anytime discrete-search contact planner described in \cite{traversability_humanoids} with substantial modification on the edge cost and heuristic computation. We formulate the contact planning problem as a graph search. Each state consists of a CoM position, a CoM velocity, and a stance represented as a set of contacting end-effector poses. An action is either moving one end-effector to a new contact pose, or breaking one palm contact. The contact transitions are based on a predefined discrete transition model, shown in Figure \ref{step_transition_model}, and we adopt the contact projection scheme in \cite{traversability_humanoids}. The edge cost of each action from a state $s$ to a state $s'$ is defined as
\begin{equation}
\Delta g(s,s') = d(s,s') + w_{s} + w_{dyn}d_{dyn}(s,s')
\label{edge_cost}
\end{equation}
\noindent where $d(s,s')$ is the XY distance the contact end-effectors' mean position travels in the contact transition, $w_{s} \in \mathbb{R^{+}}$ is a fixed cost of a contact transition, $d_{dyn}$ is the dynamics cost, which captures the dynamical robustness of the contact transition. The dynamic cost is the optimal objective value of the dynamics optimization, discussed in Sec. \ref{com_dynopt}, for the contact transition. Running the optimization in the planner is too time consuming, and we will describe how to estimate such a cost in Sec. \ref{learn_the_results_of_dynopt}. $w_{dyn} \in \mathbb{R^{+}}$ captures how much emphasis a user wants to put on minimizing the total dynamics cost of the path. In practice, robust contact sequences may contain more steps, and the user can adjust $w_{dyn}$ to trade-off between the number of steps and dynamic robustness.

We solve the contact planning problem with Anytime Non-parametric A*(ANA*) algorithm \cite{ana_star}. ANA* is an anytime variation of the A* algorithm. It first performs depth-first-like search, and improves the solution over time. In this way, the robot can quickly have a feasible solution when the available planning time is limited, and get an improved solution if there is time to spare.


To guide the ANA* search, we define the heuristic function by computing the distance to reach the goal with a simplified robot model, a floating box traveling on an SE(2) grid. We use an 8-connected grid transition model, and prune out cells where there is a collision between the box and the environment. We then plan on this grid from the cell containing the goal to every other cell in the environment using Dijkstra's algorithm. The result is a policy giving a motion direction for every cell, which can also be used to estimate the amount of motion needed to reach the goal, which we terms $d_{Dijkstra}(s)$. During contact planning, the planner queries this policy with the contacting end-effectors' weighted mean position on the XY plane, and the mean feet rotation about Z axis to compute the heuristic
%
\begin{equation}
h(s) = d_{Dijkstra}(s) + w_{s}\frac{d_{Dijkstra}(s)}{\Delta d_{max}}
\label{heuristics}
\end{equation}
%
\noindent where $\Delta d_{max}$ is an overestimate of the maximum length the weighted mean contact pose can travel in one transition. The above heuristic is an example implementation for our application. It can be swapped with other heuristics, such as a Euclidean distance heuristic, or a simplified robot model policy in a discretized SE(3) space, depending on the application.

The heuristic function in Eq. \ref{heuristics} depends on the distance of the current contact poses to the goal, and does not contain any information about future dynamics cost. While ANA* will improve the solution over time, the time needed to improve the solution relies on the accuracy of the heuristic estimating future cost. The planner may be stuck in a cul-de-sac, and can only escape when the states in the cul-de-sac are exhausted. Since ANA* behaves like a depth-first search in the beginning, a cul-de-sac is especially hard for it to escape.

To ease the problem, we adopt the $\epsilon$-greedy strategy \cite{epsilon_greedy_bfs}: With probability $1-\epsilon \ (0\leq\epsilon<1)$, the planner expands a node using the same rule as ANA*, and with probability $\epsilon$, it randomly explores a node in the priority queue. Since the random exploration does not prune out any nodes in the priority queue, and can only find new nodes or lower-cost paths to reach existing nodes, this variation does not affect the guarantees of ANA*. This strategy helps the planner escape cul-de-sacs faster by enabling the planner to explore nodes outside the cul-de-sac before exhausting it.

\begin{figure}
\subfloat{
    \adjustbox{width=0.72\columnwidth,valign=B}{
    \begin{tabular}{|c|l|l|c|}
    \hline
    \multicolumn{1}{|l|}{Index} & \multicolumn{1}{c|}{\begin{tabular}[c]{@{}c@{}}Initial\\ Contacts\end{tabular}} & \multicolumn{1}{c|}{Contact Transition} & Dim. \\ \hline
    0 & \multirow{2}{*}{\begin{tabular}[c]{@{}l@{}}Only foot\\ contacts\end{tabular}} & Move a foot contact & 24 \\ \cline{1-1} \cline{3-4} 
    1 &  & Add a palm contact & 24 \\ \hline
    2 & \multirow{5}{*}{\begin{tabular}[c]{@{}l@{}}Foot contacts\\ and a palm\\ contact\end{tabular}} & Move the inner foot contact & 30 \\ \cline{1-1} \cline{3-4} 
    3 &  & Move the outer foot contact & 30 \\ \cline{1-1} \cline{3-4} 
    4 &  & Break the palm contact & 24 \\ \cline{1-1} \cline{3-4} 
    5 &  & Move the palm contact & 30 \\ \cline{1-1} \cline{3-4} 
    6 &  & Add the other palm contact & 30 \\ \hline
    7 & \multirow{3}{*}{\begin{tabular}[c]{@{}l@{}}All foot and\\ palm contacts\end{tabular}} & Move a foot contact & 36 \\ \cline{1-1} \cline{3-4} 
    8 &  & Break a palm contact & 30 \\ \cline{1-1} \cline{3-4} 
    9 &  & Move a palm contact & 36 \\ \hline
    \end{tabular}}
}
\subfloat{
    \includegraphics[width=0.23\linewidth]{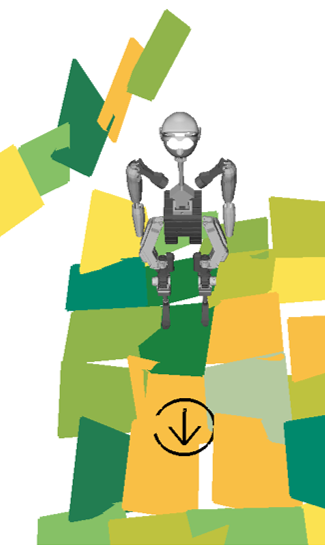}
}
\caption{Left: All categories of the contact transitions. The inner or outer foot means the foot in the same or opposite side of the palm contact. Each dimension includes all the initial contact poses, the new contact pose (if there is any), and initial CoM position and velocity. Right: An example environment to collect the training data. The tilting angle of each surface, the wall orientation, and wall distance to the robot are randomly sampled.}
\label{contact_transition_category}
\end{figure}

\section{Evaluation of the Dynamics of Contact Transitions}

To precisely evaluate the dynamics cost $d_{dyn}$ of a contact transition to a new state, a dynamics optimization from the initial state to the new state is required. However, it is not only time consuming to compute, but also difficult to learn because the input dimension can be arbitrarily high depending on the depth of the new state in the search tree. Therefore, we approximate the dynamics evaluation as only the dynamics optimization of the contact transition. Only after the contact sequence is returned by the planner, we then apply dynamics optimization on the whole contact sequence to finally output the dynamics sequence. However, even with this simplification, running dynamics optimization for every contact transition in a search tree is still too time consuming (in the order of 100 ms) for practical use. Therefore, we propose to learn the prediction of the results of the dynamics optimization of each contact transition using neural networks. In our test, each query to the network takes about 0.1 ms, which is 3 orders of magnitude faster than the original dynamics optimization.

\section{Learning the Result of the Dynamics Optimization of Contact Transitions} \label{learn_the_results_of_dynopt}


For each contact transition, the dynamics optimizer needs to decide if it is dynamically feasible, compute the objective value as part of the edge cost, and output the CoM position and velocity of the child state. To capture the function of the dynamics optimizer in contact planning, we train two kinds of neural networks:

\begin{itemize}
  \item A classifier to predict the dynamic feasibility
  \item A regressor to estimate the objective value, and the CoM position and velocity after the contact transition
\end{itemize}

\noindent The classifier has 1D binary output, which represents the feasibility of the transition, and the regressor has 7D continuous value outputs, which includes 1D objective, 3D CoM position, and 3D CoM velocity. The inputs of the neural networks are all the contact poses in the contact transition, and the initial CoM position and velocity, as same as the dynamics optimizer. To simplify the problem , we ignore CoM angular velocities in the input/output vectors, and encode the angular momentum in the objective function. We train separate neural network for each kind of contact transition using different end-effectors. Since most of the humanoid robots have symmetric kinematic structure, we further exploit this symmetry to define 10 categories of contact transition, and show its corresponding input dimensions in Figure \ref{contact_transition_category}.

The training data are collected by running the planner which calls the dynamics optimizer in each new branch in randomly tilted surface environments, as shown in Figure \ref{contact_transition_category}. The environments allow us to collect contact transitions with various contact locations and orientations. Each contact pose is encoded as a $\mathbb{R}^{6}$ vector with position and orientation in Tait-Bryan angles. Each angle is set to be in $[-\pi, \pi)$ to avoid the confusion of other coterminal angles. To capture the spatial relationship of the orientation data which contain angles near $\pi$ and $-\pi$, we duplicate those samples with $\pm2\pi$ in the training data, but always query the neural network with angles within $[-\pi, \pi)$.

\begin{figure}[t]
	\centering
	\medskip
	\includegraphics[scale=0.64]{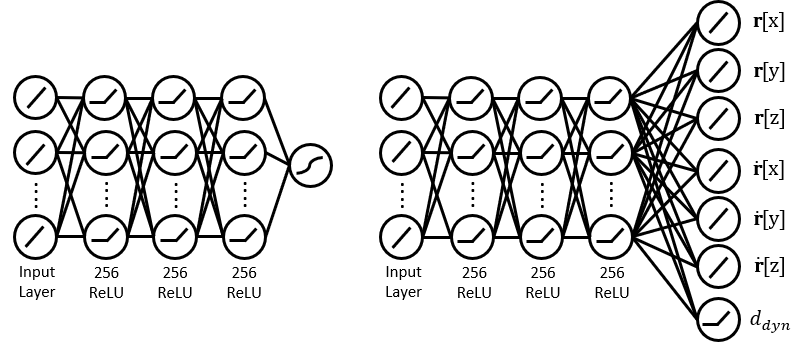}
	\vspace{-0.25in}
	\caption{Left: the classification network. Right: the regression network.}
	\label{neural_network}
\end{figure}

The neural networks used in this work are shown in Figure \ref{neural_network}.
Although it is possible to find the best-performing network structure for each category of contact transitions, we find out that using the same structure for all categories performs reasonably well, and is much simpler in implementation. For the classifier network, the output layer uses softmax activation function, which makes the network a logistic regressor. For the regression network, the output layer is a combination of linear functions for CoM position and velocity, and ReLU for the objective value. ReLU ensures the network to output positive objective values. The hidden layers for both networks are the same, which are 3 layers of 256 fully-connected nodes using ReLU activation function.


\begin{figure}[t]
	\centering
	\smallskip
	\includegraphics[scale=0.35]{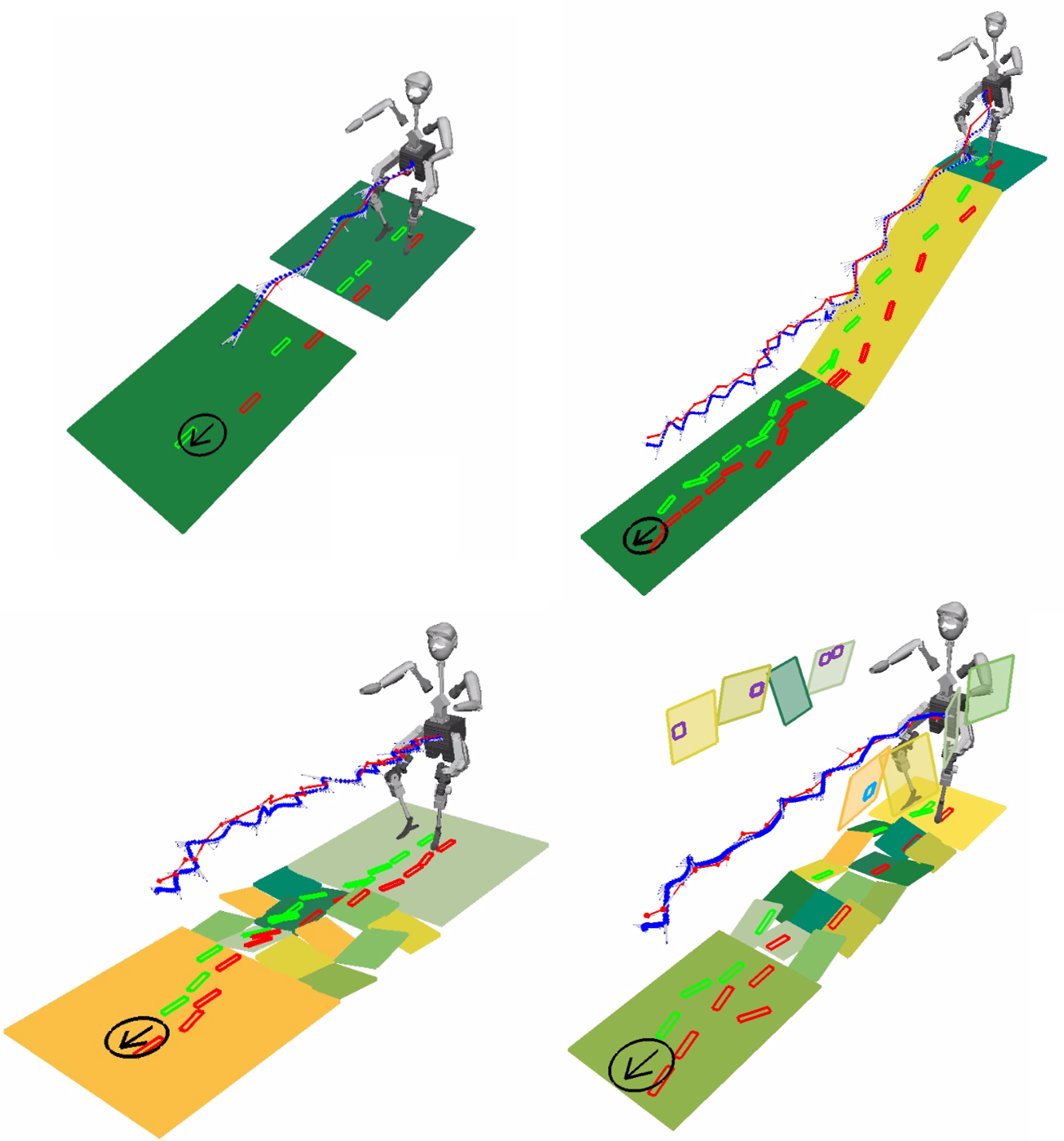}
	\vspace{-0.1in}
	\caption{Planning examples of the proposed approach for wide gap (top left), steep slope (top right), rubble field (bottom left) and rubble corridor (bottom right) environments. The red line and blue line mark the predicted CoM trajectory, and the CoM trajectory returned by the dynamics optimizer, respectively. Contact sequences include left foot(red), right foot(green), left palm(cyan), and right palm(magenta) contacts.}
	\vspace{-0.15in}
	\label{planning_example}
\end{figure}

\section{Experiments}

We evaluate the performance of the proposed approach in four environments in simulation: a wide gap, a steep slope, a rubble field, and a rubble corridor, as shown in Figure \ref{planning_example}. For each test, we set $w_{s}=3$, $w_{dyn}=0.1$, $\epsilon=0.1$, and 30 seconds time limit for the proposed approach. The contact planner will keep improving solutions within this time limit, and outputs all solutions during the improvement process. With all the contact sequences returned by the ANA*, we run a complete dynamics optimization to generate a full motion sequence, from the latest to the first contact sequence until a dynamically feasible one is confirmed. For all the dynamics optimization, we fix the time step to be 0.2 second. We also fix the timing for each contact transition: 1 second in original contact (shifting CoM) and 1 second for moving the end-effector. The friction coefficient is 0.5. The weights of each term in the objective function are: $\mathbf{l}$:0.2, $\mathbf{\dot{l}}$:0.01, $\mathbf{k}$:1, $\mathbf{\dot{k}}$:0.3, $\mathbf{f}_{e}$:0.01, $\tau_{e}$:1, $\mathbf{f}_{l}$:10, and $\mathbf{z}_{w}$:1. \textcolor{black}{All parameters are chosen empirically to help generate kinodynamically feasible motion.} The dynamics optimization used in our approach is solved using the Ipopt solver \cite{ipopt}. The neural networks are trained offline with Keras 2.1.6 \cite{keras} with Tensorflow 1.10.1 backend \cite{tensorflow} for 100 training epochs, and are queried online using frugally-deep \cite{frugally_deep}. All experiments were run on an Intel i7-6700 8-core 3.4GHz CPU. The proposed approach only uses a single thread. \textcolor{black}{The robot has 30 DOF; 7 DOF in each manipulator and 2 torso DOF. We show the generated trajectories in the visualizer provided by the SL simulator \cite{sl} in the attached video.}

\begin{figure}[t]
	\centering
	\includegraphics[scale=0.35]{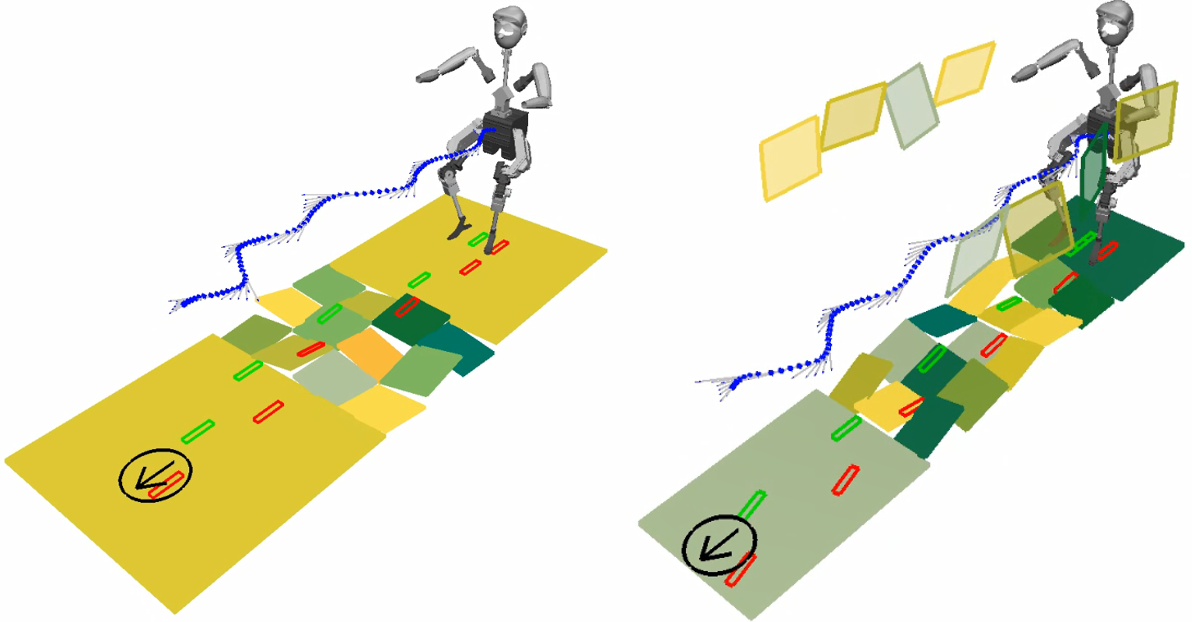}
	\vspace{-0.1in}
	\caption{Planning examples of the quasi-static contact planner for rubble field (left) and rubble corridor (right) environments.}
	\label{quasi_static_planning_example}
\end{figure}

\begin{figure}[t]
\medskip
\subfloat{
\adjustbox{width=0.97\columnwidth,valign=t}{
\footnotesize
\begin{tabular}{|c|c|c|c|}
\hline
\multirow{2}{*}{\begin{tabular}[c]{@{}c@{}}The Proposed \\ Approach\end{tabular}} & \multicolumn{3}{c|}{\begin{tabular}[c]{@{}c@{}}Mixed integer contact planner \\ with simplified dynamics model\end{tabular}} \\ \cline{2-4} 
 & 12 Contacts & 18 Contacts & 24 Contacts \\ \hline
\textbf{0.098 $\pm$ 0.037} & 85.93 $\pm$ 56.41 & 33.93 $\pm$ 18.54 & 46.40 $\pm$ 20.30 \\ \hline
\end{tabular}}}
\caption{Time required to find dynamically feasible contact sequence in rubble field environments (Unit: second)}
\label{rubble_field_MIP_table}
\end{figure}

In the following experiments, we compare our approach with a baseline quasi-static contact planner, which tries to find the shortest quasi-static contact sequence to the goal. The quasi-static contact planner follows the formulation shown in Section \ref{anytime_contact_planner}, but it does not consider any dynamics, and only verifies the static balance of the robot stance at each state using \cite{giwc}. We also impose the 30-second time limit, and use dynamics optimization on the contact sequence generated by the quasi-static contact planner to find its dynamics sequence. In addition to the quasi-static contact planner, we also compare the proposed planner with a mixed-integer contact planner \cite{centroidal_dynopt_1} in the rubble field environment to show the advantage of the proposed approach in a non-trivial environment.

\subsection{Wide Gap Environment Test}

In this test, we show that the proposed approach can plan dynamically feasible contact sequence to cross a 0.5 meter wide gap on the ground. Including the length of the robot feet, the robot has to make a 0.72 meter stride to cross the gap, which is impossible to achieve by quasi-static walking. We use a dedicated foot contact transition model for making large step in this test, but query the same neural networks. Figure \ref{planning_example} shows the contact plan, and CoM trajectory returned by the proposed approach. It took 0.143 seconds to find the contact sequence, and 1.23 seconds for dynamics optimization over the contact sequence.

\begin{figure*}[t]
\centering
\medskip
\subfloat{
\footnotesize
\begin{tabular}{|c|c|c|c|c|c|c|c|c|c|c|c|c|c|}
\hline
Test & Approach & (1) & (2) & (3) & (4) & (5) & (6) & (7) & (8) & (9) & (10) & (11) & (12) \\ \hline
\multirow{2}{*}{\begin{tabular}[c]{@{}c@{}}Rubble Field\\ Environment\end{tabular}} & \begin{tabular}[c]{@{}c@{}}Quasi-static \\ Contact Planner\end{tabular} & 47/50 & 47/47 & 1.17 & 3498.5 & 12.0 & 12.7 & \textbf{0.472} & 0.706 & 0.767 & 0.012 & \textbf{0.062} & 0.0068 \\ \cline{2-14} 
 & \begin{tabular}[c]{@{}c@{}}The Proposed \\ Approach\end{tabular} & \textbf{50/50} & 50/50 & \textbf{1.02} & \textbf{3334.2} & \textbf{6.00} & \textbf{9.68} & 0.581 & \textbf{0.559} & \textbf{0.763} & \textbf{0.010} & 0.079 & \textbf{0.0079} \\ \thickhline
\multirow{2}{*}{\begin{tabular}[c]{@{}c@{}}Rubble Corridor\\ Environment\end{tabular}} & \begin{tabular}[c]{@{}c@{}}Quasi-static \\ Contact Planner\end{tabular} & 44/50 & 44/44 & \textbf{1.05} & 3418.7 & 11.5 & 11.16 & \textbf{0.523} & 0.618 & 0.768 & \textbf{0.013} & \textbf{0.082} & 0.0069 \\ \cline{2-14} 
 & \begin{tabular}[c]{@{}c@{}}The Proposed \\ Approach\end{tabular} & \textbf{50/50} & 49/50 & 1.59 & \textbf{2392.5} & \textbf{4.38} & \textbf{4.83} & 1.378 & \textbf{0.349} & \textbf{0.631} & 0.025 & 0.088 & \textbf{0.0173} \\ \hline
\end{tabular}}
\caption{Results for the rubble field corridor environments: (1) Contact planning and (2) Dynamics optimization success rates (3) Average number of tested contact sequence to find a dynamically feasible sequence (4) Mean dynamics objective of the whole contact sequence (5) Mean lin. momentum norm (kg$\cdot$m/s) (6) Mean lin. momentum rate norm (kg$\cdot$m/s$^{2}$) (7) Mean angular momentum norm (kg$\cdot$m$^{2}$/s) (8) Mean angular momentum rate norm (N$\cdot$m) (9) Mean RMS contact force norm (10) Mean contact torque (N$\cdot$m) (11) Mean lateral contact force norm (12) Mean CoP distance to contact boundary (m). Contact forces are normalized by the robot weight and are unitless. In (5)-(12), means are computed over all time steps of all dynamically feasible trials.}
\label{quantitative_comparison}
\vspace{-0.1in}
\end{figure*}

\subsection{Steep Slope Environment Test}

In this test, the robot is required to go down a 3 meter long $30^{\circ}$ slope. The robot cannot maintain static balance on the slope, so the quasi-static contact planner is not able to find any solution. Our approach finds the first solution in 0.702s, and generating the contact sequence shown in Figure \ref{planning_example} takes 10.617s. The dynamics optimization takes 6.32s to generate the dynamics sequence which contains 31 contacts.

\subsection{Rubble Field Environment Test} \label{rubble_field_env_test}
The rubble field environment (Fig. \ref{planning_example}), simulates a common disaster-relief scenario. The robot dynamically walks over a rubble to reach a goal about 3.4 meter away. Contact surfaces are randomly tilted in X and Y axes in $[-20^{\circ}, 20^{\circ}]$. The environment contains 14 convex contact surfaces.

In this test, we compare the performance of the proposed approach with the mixed integer contact planners. We first compared to a custom implementation of a mixed integer contact planner that internally solves the dynamics optimization problem  as in \cite{centroidal_dynopt_2}, which is also used for training our neural networks. After 7 hours of planning time, it was not able to find a feasible solution. We then used a simplified dynamics model \cite{centroidal_dynopt_1} and assumed that the contacts are all point contacts, which fixes each CoP to one point, and neglects the contact orientations. We solved it with state-of-the-art mixed integer solver, Gurobi 8.0 \cite{gurobi}, using 8 threads. As shown in Figure \ref{rubble_field_MIP_table}, the mixed integer contact planner using the simplified model still takes much longer than the proposed approach to find a feasible solution. Furthermore, the mixed integer contact planner requires the user to specify the number of contacts used in the plan. Since the planning is in unstructured environments, it is not trivial to decide how many contacts are needed, and different number of contacts can have a great impact on the planning time (Figure \ref{rubble_field_MIP_table}).

Compared to the quasi-static contact planner, the proposed approach produces contact sequences with similar dynamics objective. However, as shown in Figure \ref{quantitative_comparison}, the proposed approach generates motion with lower linear momentum and rates of linear and angular momenta. The angular momentum of the proposed approach is higher because it does not always produce straight walking motion as the quasi-static contact planner normally does, instead it may take a detour to achieve more robust locomotion using our approach.


\subsection{Rubble Corridor Environment Test}

In this test, we set up the rubble corridor environment, where palm contacts are available, and test the planner's ability to find dynamically robust contact sequence in such environment. The surfaces are randomly tilted as in Section \ref{rubble_field_env_test}. Without any user specification, the proposed approach is able to discover palm contacts in the search, as shown in Figure \ref{planning_example}. The quasi-static contact planner, on the other hand, does not consider the dynamics, and favors path with shorter traveling distance and fewer number of contacts. Therefore, it outputs solutions without palm contact, as shown in Figure \ref{quasi_static_planning_example}. Compared to the quasi-static contact planner, the proposed approach generates motion with lower linear momentum, rates of the linear and angular momenta, and higher CoP clearance to the contact boundary, as shown in Figure \ref{quantitative_comparison}. Although the angular momentum of the motion generated by the propose approach is much higher, the robot momenta rates are much lower, which results in  a much lower dynamics objective of the whole contact sequence.

\subsection{Prediction of Dynamics Optimizer Results}

\begin{figure}[t]
	\centering
	\includegraphics[scale=0.27]{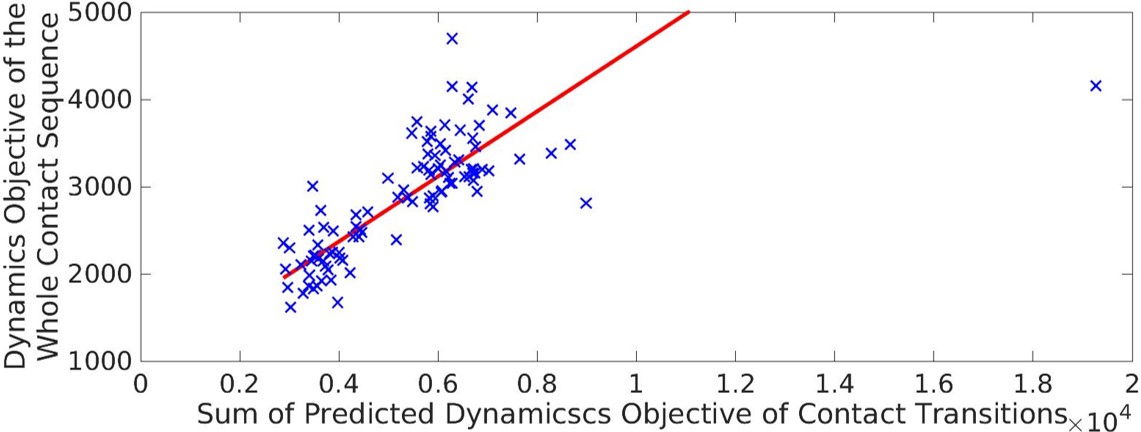}
	\caption{Relationship between the sum of the predicted dynamics objective of contact transitions and the actual dynamics objective of the whole contact sequence. Data taken from the rubble field and rubble corridor environments. \textcolor{black}{The linear model showing the correlation is fit with robust regression \cite{robust_regression}.}}
	\label{dynamics_objective_relationship}
\end{figure}

Here, we analyze the performance of the neural network in predicting useful information to guide the planner to find dynamically robust contact sequences. Figure \ref{nn_model_performance} summarizes the networks' performance on predicting the results of the dynamics optimization over each contact transition. For each motion category, we use $10^{5}$ training data, and tested with another $1000$ data. The proposed approach estimates the dynamics objective of the whole contact sequence with the sum of dynamics objective in each contact transition of the contact sequence. As shown in Figure \ref{dynamics_objective_relationship}, this estimates is not accurate as it neglects previous and later contact poses in each optimization over a contact transition. However, the estimates and the actual dynamics objective are highly correlated, which makes the estimates a suitable edge cost function to select branches which lead to lower dynamics objective of the whole contact sequence.

\begin{figure}
\centering
\subfloat{    
    \scriptsize
    \begin{tabular}{|c|c|c|c|c|c|}
    \hline
    \multirow{2}{*}{\begin{tabular}[c]{@{}c@{}}Contact\\ Transition\\ Category\\ Index\end{tabular}} & \multirow{2}{*}{\begin{tabular}[c]{@{}c@{}}Dynamic\\ Feasibility\\ Prediction\\ Accuracy\end{tabular}} & \multirow{2}{*}{\begin{tabular}[c]{@{}c@{}}Mean\\ Actual\\ Dynamics\\ Objective\end{tabular}} & \multicolumn{3}{c|}{\begin{tabular}[c]{@{}c@{}}Mean Absolute \\ Error in Regression\end{tabular}} \\ \cline{4-6} 
     &  &  & \begin{tabular}[c]{@{}c@{}}Dynamics\\ Objective\end{tabular} & \begin{tabular}[c]{@{}c@{}}Final\\ CoM\\ (mm)\end{tabular} & \begin{tabular}[c]{@{}c@{}}Final CoM\\ Velocity\\ (mm/s)\end{tabular} \\ \hline
    1 & 90.3\% & 1436.10 & 62.45 & 7.5 & 6.6 \\ \hline
    2 & 97.0\% & 740.85 & 40.38 & 6.0 & 5.4 \\ \hline
    3 & 95.3\% & 164.96 & 20.70 & 9.0 & 5.4 \\ \hline
    4 & 93.5\% & 119.85 & 11.07 & 6.7 & 4.7 \\ \hline
    5 & 94.3\% & 516.53 & 45.06 & 7.1 & 4.1 \\ \hline
    6 & 95.2\% & 87.80 & 12.39 & 9.1 & 4.1 \\ \hline
    7 & 98.1\% & 53.47 & 8.10 & 7.3 & 4.1 \\ \hline
    8 & 96.6\% & 50.66 & 17.28 & 8.1 & 2.4 \\ \hline
    9 & 96.1\% & 88.00 & 15.18 & 9.0 & 3.0 \\ \hline
    10 & 98.3\% & 62.40 & 7.56 & 8.1 & 3.7 \\ \hline
    \end{tabular}
}
\caption{Performance of the neural networks to predict dynamic feasibility, dynamics objective, final CoM and CoM velocity of a contact transition. Refer to Figure \ref{contact_transition_category} for the meaning of each contact transition category index.}
\label{nn_model_performance}
\end{figure}

\section{Conclusion}
We proposed a contact planner which finds dynamically robust contact sequence involving both foot and palm contacts. Costly dynamics optimization is replaced by a learned
prediction of dynamic feasibility and edge cost. The planner can leverage these learned functions to efficiently evaluate contact options in the planning loop. In the future, we would like to extend the contact planner to further consider timing of each contact transition \cite{centroidal_dynopt_2}, so that the contact planner can generate a wider variety of dynamic motions.







\bibliography{references}{}
\bibliographystyle{unsrt}

\end{document}